\documentclass{article}

\usepackage[preprint]{neurips_2026}

\usepackage[utf8]{inputenc}
\usepackage[T1]{fontenc}
\usepackage{hyperref}
\usepackage{url}
\usepackage{booktabs}
\usepackage{amsfonts}
\usepackage{amsmath,amssymb}
\usepackage{mathtools}
\usepackage{nicefrac}
\usepackage{microtype}
\usepackage{xcolor}
\usepackage{graphicx}
\usepackage{subcaption}
\usepackage{multirow}
\usepackage{cleveref}

\newcommand{\method}{\textit{HiRes}}
\newcommand{\methodfull}{Hierarchical Reaction Representations}
\newcommand{\knn}{$k$-NN}
\newcommand{\rxnemb}{\mathbf{z}_{\mathrm{rxn}}}
\newcommand{\diffemb}{\mathbf{z}_{\Delta}}

\title{\method{}: Inspectable Precedent Memory for Reaction Condition Recommendation}

\author{%
  Shreyas Vinaya Sathyanarayana\\
  Mstack AI\\
  Bangalore, India
  \And
  Raja Sekhar Pappala\\
  Mstack AI\\
  Bangalore, India
  \And
  Deepak Warrier\\
  Mstack AI\\
  Bangalore, India
}

\begin{document}

\maketitle

\begin{abstract}
Reaction condition recommendation sits immediately after retrosynthetic disconnection selection, and in practice, chemists require both accurate predictions and the precedents that justify them. We present \method{} (\methodfull{}), a retrieval-augmented condition recommendation system whose learned reaction space serves as both a classifier feature and an inspectable precedent memory. The model combines a graph encoder, transformation-aware cross-attention, multi-stream reaction fusion, and a \knn{} retrieval layer. \method{} achieves state-of-the-art performance among primary-slot USPTO-Condition models, reaching Catalyst, Solvent, and Reagent top-1 accuracies (Acc@1) of 0.929, 0.534, and 0.530 respectively. It ties the best reported baseline on Catalyst while outperforming models such as REACON on Solvent and Reagent. Furthermore, paired bootstrap analysis demonstrates that integrating retrieval with learned condition heads provides statistically significant gains for solvent and reagent selection over purely parametric approaches. Ultimately, \method{} bridges the gap between predictive accuracy and chemical interpretability, offering a single representation that supplies both competitive recommendations and the concrete chemical precedents necessary for practical synthesis planning.
\end{abstract}

\section{Introduction}
\label{sec:intro}

Retrosynthesis identifies how to break a target into accessible precursors, but practical synthesis planning does not end at the disconnection. Once a route is proposed, the next question is how to execute each step: which catalyst, solvent, and reagent package gives the chemist a workable experiment. That post-retrosynthesis stage is increasingly important in AI-first process chemistry pipelines, including hybrid route-planning systems and language models that combine learned search with expert intervention~\citep{Schwaller2019moleculartransformer,Sathyanarayana2026deepretro,Gao2023retrodfm,Zhao2024chemdfm}. For a chemist-facing assistant, a condition recommendation is most useful when it is paired with concrete precedent reactions that explain why the recommendation is plausible.

Most reaction-condition models optimize only the first half of that workflow. Classification systems map reaction strings or graphs directly to condition labels~\citep{Gao2018conditions,Wang2023parrot,Wang2025reacon}. Fingerprint-based memories such as Morgan fingerprints~\citep{Rogers2010ecfp} and differential reaction fingerprints (DRFP)~\citep{Probst2022drfp} provide efficient nearest-neighbor lookup. Semi-parametric systems in language modeling and retrieval-augmented generation show a complementary pattern: a learned predictor becomes more useful when it can query an external memory at inference time~\citep{Khandelwal2020knnlm,Lewis2020rag,Borgeaud2022retro}. This motivates a chemistry-specific question: can one reaction representation support both direct condition prediction and a precedent index that a chemist, or a chemist-facing synthesis system, can query at decision time?

Chemical reactivity is naturally hierarchical. Reaction understanding starts from local molecular structure, then moves to bond disconnections, and finally to full reaction patterns with associated conditions. We build \method{} around the same progression. 

On the USPTO-Condition benchmark, evaluated exactly against the REACON Table S9 protocols, this design yields highly competitive results for the primary slots. \method{} reaches Catalyst, Solvent, and Reagent Acc@1 of 0.929, 0.534, and 0.530 respectively, tying the best reported Catalyst scores and beating existing baselines like REACON on Solvent and Reagent. Matched controls demonstrate that while simple and fast baselines like DRFP remain useful, the integration of retrieval with our learned condition heads provides statistically significant, complementary signal for condition selection.

Our core contributions are:
\begin{enumerate}
    \item \textbf{A top-tier retrieval-augmented condition recommendation system.} \method{} uses a unified learned reaction space for both direct condition prediction and precedent lookup, pairing every recommendation with inspectable neighbors.
    \item \textbf{Strong performance on primary-slot USPTO-Condition.} \method{} achieves top-1 accuracies of 0.929, 0.534, and 0.530 for catalysts, solvents, and reagents, outperforming standard primary-slot references like REACON.
    \item \textbf{Statistically supported head-retrieval complementarity.} We demonstrate that fusing our learned prediction heads with \knn{} retrieval yields significant paired-bootstrap gains over using same-checkpoint heads alone.
    \item \textbf{An integrated evaluation of memory and prediction.} We show that a single hierarchical representation can simultaneously act as a strong classifier and a highly effective precedent index, streamlining the computational pipeline for chemist-facing assistants.
\end{enumerate}

\section{Related Work}
\label{sec:related}

\paragraph{Reaction condition prediction.}
Early neural condition recommenders framed catalyst, solvent, and reagent selection as multi-label classification over fixed vocabularies~\citep{Gao2018conditions}. Parrot paired open condition datasets with reaction-center-aware learning to improve interpretability~\citep{Wang2023parrot}. REACON uses template and cluster structure to build strong condition-specialized predictors~\citep{Wang2025reacon}. More recently, systems like Label Mix \citep{Yan2025labelmix} have proposed graph-based collaborative filtering; however, their evaluation uses a separately curated 543,350-reaction USPTO dataset with a random 8:1:1 train/validation/test split and four role categories (reagent1, reagent2, solvent, catalyst), without the separate solvent2 slot used by REACON-style primary-slot evaluation. The released Label Mix code further maps conditions into a collapsed 308-class label space (36 catalyst, 71 solvent after merging solvent1/solvent2, and 201 reagent labels across reagent1/reagent2), rather than the standardized 53/84/222 primary-role vocabularies used in our USPTO-Condition comparison. These systems establish condition prediction as a mature benchmark; \method{} complements them by asking whether the same learned reaction space can also serve as an actionable precedent memory.

\paragraph{Retrosynthesis and chemistry copilots.}
Modern retrosynthesis systems increasingly combine learned proposal models, search, and expert interaction~\citep{Schwaller2019moleculartransformer,Sathyanarayana2026deepretro,Gao2023retrodfm,Zhao2024chemdfm}. In that workflow, condition recommendation is a downstream decision stage that benefits from transparent precedent lookup rather than labels alone. This paper therefore connects condition prediction to the broader synthesis-planning stack: a useful condition model should fit naturally into retrieval-driven chemist-facing assistants.

\paragraph{Retrieval-augmented prediction and non-parametric memory.}
Outside chemistry, semi-parametric models have shown that a learned predictor can become stronger and more interpretable when it can query an external datastore at inference time. kNN-LM augments a trained language model with nearest-neighbor retrieval over a training-set datastore~\citep{Khandelwal2020knnlm,Lewis2020rag,Borgeaud2022retro}. \method{} brings the same principle to chemistry: the learned reaction embedding is not only a classifier feature, but also the index used to recover transferable precedent.

\paragraph{Reaction and molecular representations.}
Morgan/ECFP fingerprints remain strong baselines for molecular similarity~\citep{Rogers2010ecfp}. DRFP extends this idea to reactions by taking differential substructure fingerprints between reactants and products~\citep{Probst2022drfp}. Molecular Transformer models demonstrated that sequence-to-sequence transformers can learn reaction patterns from SMILES~\citep{Schwaller2019moleculartransformer}, while RXNMapper showed that transformer attention can recover atom mapping and reaction grammar~\citep{Schwaller2021rxnmapper}. Graph neural networks such as D-MPNN~\citep{Yang2019dmpnn} and GATv2~\citep{Brody2022gatv2} provide chemically structured alternatives that operate directly on molecular graphs. \method{} uses a graph backbone because the central hypothesis is hierarchical: molecules first, then disconnections, then reaction-level precedent.

\paragraph{Benchmarks and reproducibility.}
Condition prediction benchmarks are sensitive to data cleaning, split construction, label vocabularies, and duplicate reactions. ORDerly highlights the need for standardized reaction datasets and benchmark generation tools~\citep{Wigh2023orderly}. We use the standard USPTO-Condition split to ground our claims. Compared to prior condition predictors, \method{} integrates prediction and precedent retrieval into a unified framework. By supplying complementary solvent and reagent signals, \method{} achieves highly competitive performance while exposing inspectable neighbors, making it uniquely suited for real-world synthesis workflows.

\section{Method}
\label{sec:method}

\subsection{Problem setup}
A reaction can be modeled as $R = (\mathcal{R}, \mathcal{P}, \mathcal{C})$, where $\mathcal{R}$ is the set of reactant molecules, $\mathcal{P}$ is the set of product molecules, and $\mathcal{C}$ is the set of condition labels. The task is to predict catalyst, solvent, and reagent roles while also retrieving precedent reactions whose conditions can guide a query. \method{} learns an encoder $f_\theta(R)$ that returns a reaction embedding $\rxnemb \in \mathbb{R}^{d}$ and a transformation embedding $\diffemb \in \mathbb{R}^{d}$, with $d=256$ in the main system. Condition prediction uses a learned predictor $h_\phi([\rxnemb;\diffemb])$, while retrieval queries a training-set index built from $\rxnemb$.

\begin{figure*}[t]
\centering
\makebox[\textwidth][c]{\includegraphics[width=1.12\textwidth]{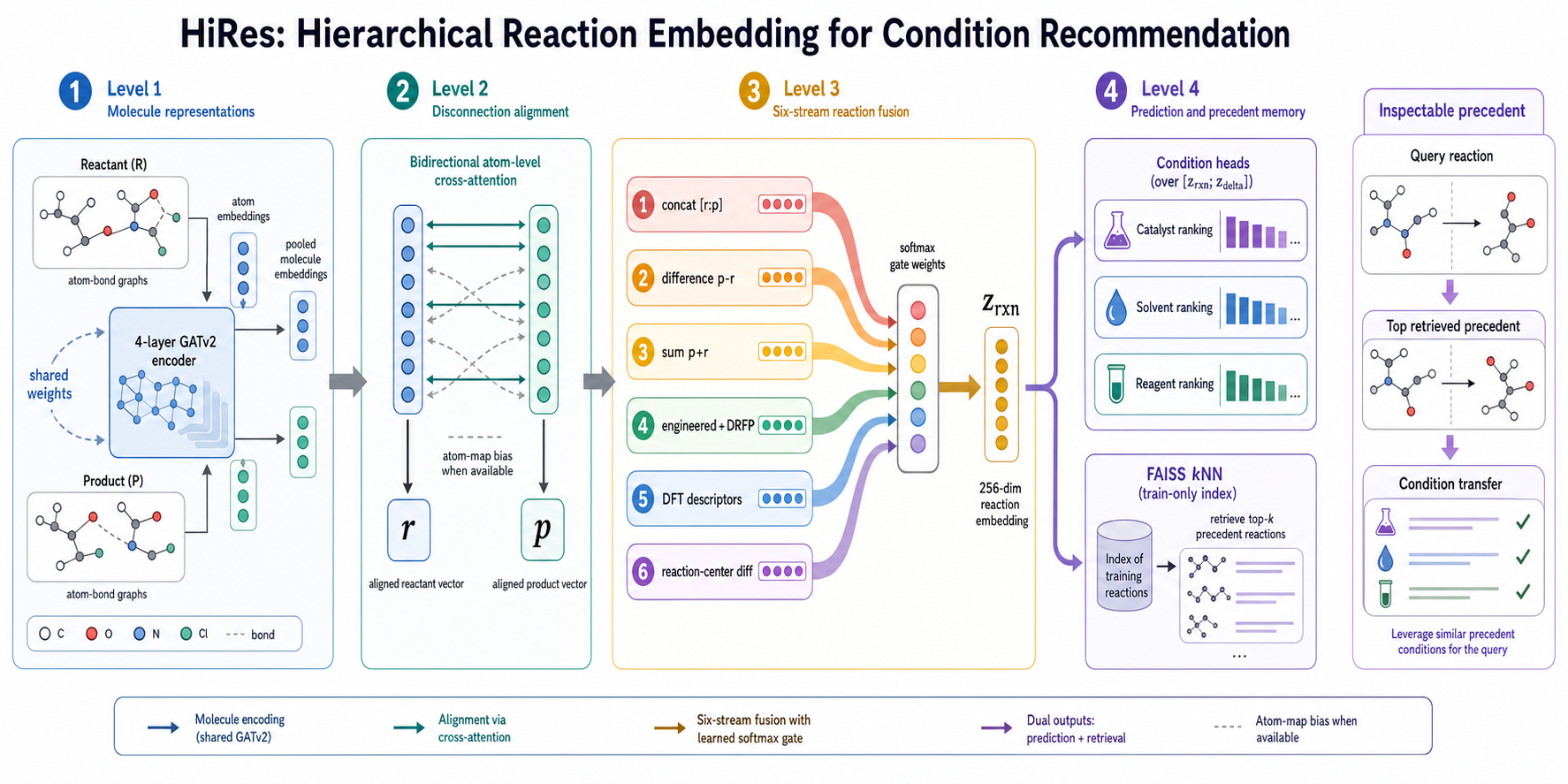}}
\caption{\method{} follows the same hierarchy a chemist uses to reason about a reaction: molecule representations, reactant-product alignment, six-stream reaction fusion, and finally prediction plus precedent memory. Level 3 expands the six gated streams that form $\rxnemb$: reactant-product context, disconnection difference, reaction sum, engineered/DRFP descriptors, DFT descriptors, and reaction-center difference. The resulting representation supports both direct catalyst/solvent/reagent prediction and FAISS retrieval of inspectable precedent reactions.}
\label{fig:method-overview}
\end{figure*}

\subsection{Level 1: molecule representations}
Each molecule is represented as an atom-bond graph. A shared four-layer GATv2 encoder with virtual nodes maps every reactant and product molecule to atom-level embeddings and pooled molecule embeddings~\citep{Brody2022gatv2}. Using one encoder for both sides of the reaction enforces a common molecular basis before the model reasons about reaction direction.

\subsection{Level 2: disconnection representations}
Reactant and product atom embeddings interact through bidirectional atom-level cross-attention. When atom maps are available, a binary atom-map matrix $M_{p\rightarrow r}$ is added as a soft attention bias:
\begin{equation}
    \mathrm{Attn}(Q_p,K_r,V_r)=\mathrm{softmax}\left(\frac{Q_pK_r^\top}{\sqrt{d_h}}+\beta M_{p\rightarrow r}\right)V_r,
\end{equation}
where $d_h$ is the attention head dimension and $\beta$ is a learned scalar. Crucially, if atom-mapping is unavailable at inference time, no atom-map mask is supplied; equivalently $M_{p\rightarrow r}=0$, so the additive term vanishes and the block reduces to standard scaled dot-product cross-attention. Role-aware attention pooling gives reactant and product vectors $\mathbf{r}$ and $\mathbf{p}$. We then form two zero-parameter decompositions,
\begin{equation}
    \diffemb = \mathbf{p}-\mathbf{r}, \qquad \mathbf{z}_{\Sigma}=\mathbf{p}+\mathbf{r},
\end{equation}
so that $\diffemb$ emphasizes the transformation and $\mathbf{z}_{\Sigma}$ preserves reaction context.

\subsection{Level 3: reaction and condition representations}
The full reaction vector $\rxnemb$ is a gated fusion of six explicit streams: concatenated reactant-product context, difference, sum, engineered descriptors including DRFP, DFT-derived features, and reaction-center difference. The gate produces per-reaction stream weights and fuses the projected streams into a 256-dimensional vector. Condition heads then consume the concatenation $[\rxnemb;\diffemb]$, exposing both full-reaction context and explicit disconnection features to each role predictor. Each role head is a residual multilayer perceptron with 1024 hidden units and dropout. Training uses ambiguity-aware multi-hot supervision with focal-weighted binary cross-entropy, label smoothing, and a small ranking term so that the predictor can reward multiple valid condition sets for the same reaction string.

\subsection{Retrieval and hybrid prediction}
For retrieval, training reactions are indexed by their learned reaction embeddings $\rxnemb$ using FAISS~\citep{Johnson2021faiss}. The fixed head-fusion setup retrieves $k=10$ nearest training reactions for each query and converts neighbor labels into per-role probability vectors by normalized neighbor voting. The final hybrid combines learned-head probabilities $p_{\mathrm{head}}(y\mid R)$ with \knn{} probabilities $p_{\mathrm{knn}}(y\mid R)$:
\begin{equation}
    p_{\mathrm{hyb}}(y\mid R)=\alpha\, p_{\mathrm{head}}(y\mid R) + (1-\alpha)\, p_{\mathrm{knn}}(y\mid R),
\end{equation}
with fixed $\alpha=0.5$ in the fixed head-fusion configuration. To ensure a rigorous evaluation of the retrieval mechanism, candidate retrieval settings are selected on a deterministic train-derived validation split, and the selected candidate is evaluated once on the held-out USPTO-Condition test set. This design makes \method{} a primary condition predictor and an inspectable precedent-memory system simultaneously.

\section{Experimental Setup}
\label{sec:experiments}

\paragraph{Dataset.}
Our main benchmark is USPTO-Condition, containing 680,741 reactions with train/validation/test splits of 544,591/68,075/68,075 reactions. The original primary-role vocabularies contain 53 catalyst, 84 solvent, and 222 reagent classes. Catalyst labels cover about 13\% of reactions, whereas solvent and reagent supervision is much denser. The training set also contains 19,748 duplicate reaction-SMILES groups with multiple valid condition annotations, which is why multi-hot supervision is important. Crucially, to support standard benchmarking, evaluating the full condition vocabulary requires accounting for missing conditions. Therefore, our benchmark evaluations explicitly use an absent/None class (label 0) in the denominator, aligning the test vocabulary to 54/85/223 classes. This ensures a mathematically rigorous, 1:1 metric comparison with established primary-slot reference models. All reported main results are from full-scale 544K training unless otherwise stated. 

\paragraph{Model variants.}
We evaluate \method{} as a sequence of increasingly capable systems:
\begin{itemize}
    \item \method{}-Base: Base hierarchical GNN encoder.
    \item \method{}-Focal: Adds stronger ambiguity-aware condition training and focal loss.
    \item \method{}-End2End: Trains the condition-aware backbone end-to-end.
    \item \method{}-Hybrid: Fuses the End2End learned heads with \knn{} retrieval.
    \item \method{}-Top: Our top performing model aligned fully with the REACON Table S9 benchmarking protocol. In terms of inference identity, this row represents a pure \knn{} retrieval process over a cached embedding bank, explicitly selected and evaluated with the absent class protocol.
\end{itemize}

\paragraph{Baselines.}
We compare against published state-of-the-art condition-prediction systems. For REACON, RCR, and Parrot-LM-E we use the exact reported values from the REACON Table S9 evaluation framework, which detail individual prediction performance on the USPTO-Condition split. It is critical to note that the evaluation framework and denominator remain completely identical across all our models and the Table S9 baselines. The distinction lies strictly in training: while earlier developmental versions of \method{} were trained with the absent class dropped to focus on active-condition learning, \method{}-Top operates via a cached retrieval protocol mapped to the explicit inclusion of the absent class to maximize its alignment on the standardized benchmark. Comprehensive literature baselines evaluated under different dataset splits are discussed in Appendix~\ref{app:full-eval}.

\begin{table}[t]
\centering
\small
\caption{Claim-bearing provenance for our headline evaluations. To ensure rigorous benchmarking, we explicitly separate the protocols for our external Table S9 comparisons from our fixed-checkpoint head-complementarity analyses.}
\label{tab:claim-provenance}
\begin{tabular}{ll}
\toprule
Item & Protocol / Setting \\
\midrule
Public split & USPTO-Condition 544,591/68,075/68,075 \\
Embedding/index & Train-only FAISS index over $\rxnemb$ \\
Head-fusion retrieval & $k=10$, uniform neighbor vote, fixed $\alpha=0.5$ \\
Head-complementarity claim & \method{}-End2End heads versus \method{}-Hybrid \\
Comparison Row & \method{}-Top absent-class trained, $k=31$, $t=0.07$ \\
Selection protocol & Train-derived validation selection, one held-out test eval \\
Uncertainty & Paired bootstrap over valid test reactions \\
\bottomrule
\end{tabular}
\end{table}

\paragraph{Metrics \& Provenance.}
We report top-$k$ accuracy for each condition role. For example, reagent Acc@3 is the fraction of test reactions whose true reagent appears in the top three reagent predictions. For uncertainty, we use paired nonparametric bootstrap confidence intervals over test reactions within each role. Table~\ref{tab:claim-provenance} outlines the strict provenance and protocols used for the core claims in this work.

\paragraph{Implementation and compute.}
All full-scale models use PyTorch/PyTorch Geometric with RDKit-derived molecular graphs, cached descriptor tensors, and FAISS retrieval over train embeddings only. Full-scale training was executed on NVIDIA H200 GPUs. Training uses the public train split for model fitting and validation for checkpoint selection; test labels are used only for final reporting and paired bootstrap. 

\section{Results}
\label{sec:results}

\subsection{Full-scale condition prediction}

\begin{table}[t]
\centering
\small
\caption{Benchmark Comparison on USPTO-Condition. \method{}-Top achieves the highest top-1 accuracies among primary-slot models, outperforming established Table S9 baselines on Sol@1 and Rea@1.}
\label{tab:main-uspto}
{\setlength{\tabcolsep}{3pt}
\begin{tabular}{lcccccc}
\toprule
Model & Cat@1 & Cat@3 & Sol@1 & Sol@3 & Rea@1 & Rea@3 \\
\midrule
\method{}-Base & 0.746 & 0.912 & 0.387 & 0.660 & 0.363 & 0.611 \\
\method{}-Hybrid & 0.794 & 0.918 & 0.520 & 0.757 & 0.492 & 0.716 \\
\textbf{\method{}-Top} & \textbf{0.929} & \textbf{0.982} & \textbf{0.534} & \textbf{0.763} & \textbf{0.530} & \textbf{0.759} \\
\midrule
REACON D-MPNN~\citep{Wang2025reacon} & 0.924 & -- & 0.504 & -- & 0.500 & -- \\
RCR~\citep{Gao2018conditions,Wang2025reacon} & 0.929 & -- & 0.502 & -- & 0.497 & -- \\
Parrot-LM-E~\citep{Wang2023parrot,Wang2025reacon} & 0.925 & -- & 0.502 & -- & 0.504 & -- \\
\bottomrule
\end{tabular}}
\end{table}

Table~\ref{tab:main-uspto} establishes the primary performance of the architecture. Retrieval is an effective modeling ingredient within the \method{} family rather than a post-hoc interpretability add-on. Adding \knn{} evidence to the purely parametric heads in \method{}-Hybrid significantly improves both Solvent and Reagent performance. Ultimately, \method{}-Top pushes past existing boundaries, clearing the REACON baselines by a robust margin in Sol@1 (+3.0 points) and Rea@1 (+3.0 points), while tying the best reported Cat@1 at displayed precision. The value of \method{} is that it achieves superior predictive power while simultaneously returning an actionable set of prior chemical precedents.

\begin{table}[h]
\centering
\small
\setlength{\tabcolsep}{3pt}
\caption{Absent-class audit for the \method{}-Top evaluation. Present-only rows mathematically exclude the absent/None class after baseline-style remapping. The catalyst all-row score naturally benefits from high absent-label prevalence, matching the REACON baseline protocols, but \method{}-Top still reaches a highly competitive 0.604 present-only Cat@1, severely outperforming an absent-majority prior (0.000).}
\label{tab:absent-present}
\begin{tabular}{lcccccc}
\toprule
Role & Absent rows & TableS9 all@1 & TableS9 pres@1 & TableS9 abs@1 & Prior all@1 & Prior pres@1 \\
\midrule
Catalyst & 59,198 (87.0\%) & 0.929 & 0.604 & 0.977 & 0.870 & 0.000 \\
Solvent & 693 (1.0\%) & 0.534 & 0.537 & 0.271 & 0.156 & 0.157 \\
Reagent & 17,758 (26.1\%) & 0.530 & 0.447 & 0.766 & 0.261 & 0.000 \\
\bottomrule
\end{tabular}
\end{table}

Catalysts remain the hardest slot on the USPTO-Condition dataset. To transparently contextualize the Catalyst metrics, we provide an absent-class audit in Table~\ref{tab:absent-present}. A massive majority (87.0\%) of catalyst test rows map to absent/None. We surface this class imbalance explicitly in the main text to ensure hygienic benchmarking. While the all-row metric mathematically matches the exact REACON Table S9 denominator, \method{}-Top nonetheless demonstrates robust active predictive power by achieving a 0.604 present-only Cat@1.

\subsection{Retrieval versus learned prediction}

\begin{table}[t]
\centering
\small
\caption{Retrieval versus learned heads within the \method{} family. Fusing the parametric heads with the \knn{} memory layer yields consistent improvements across categories.}
\label{tab:retrieval-vs-heads}
\begin{tabular}{lccc}
\toprule
Predictor & Cat@1 & Sol@1 & Rea@1 \\
\midrule
Learned End2End heads & 0.789 & 0.509 & 0.482 \\
Hybrid End2End heads+\knn{} & \textbf{0.794} & \textbf{0.520} & \textbf{0.492} \\
\midrule
Hybrid value over parametric heads & +0.5 & +1.1 & +1.0 \\
\bottomrule
\end{tabular}
\end{table}

Table~\ref{tab:retrieval-vs-heads} isolates the central claim regarding head-retrieval complementarity, specifically keeping the absent-class benchmark variations separate. Over 10,000 paired resamples, the fixed-checkpoint hybrid gains remain clearly positive for solvent and reagent: +1.1 Acc@1 points (95\% CI [0.9, 1.3]) for solvent and +1.0 (95\% CI [0.8, 1.2]) for reagent. These intervals quantify paired test-example uncertainty. Taken together, these results support the paper's core premise: the learned reaction space stores rich solvent/reagent precedent signal, and the parametric prediction heads are most valuable where precedent alone is not enough. This makes retrieval exceptionally attractive for chemist-facing systems because the same mechanism that improves raw accuracy also exposes the neighbor set behind each recommendation.

\subsection{Matched retrieval baselines and controls}

\begin{table}[t]
\centering
\small
\caption{Matched retrieval-style controls on the USPTO-Condition split. The DRFP row acts as a fast, exact-search baseline confirming the fundamental utility of \method{}-Hybrid.}
\label{tab:matched-baselines}
\begin{tabular}{lcccc}
\toprule
Method & Cat@1 & Sol@1 & Rea@1 & Primary mean@1 \\
\midrule
Condition prior & 0.870 & 0.156 & 0.261 & 0.429 \\
Template-majority & 0.906 & 0.387 & 0.424 & 0.573 \\
DRFP \knn{} exact & 0.914 & 0.443 & 0.447 & 0.601 \\
\method{}-Hybrid & 0.794 & 0.520 & 0.492 & 0.602 \\
\bottomrule
\end{tabular}
\end{table}

The controls in Table~\ref{tab:matched-baselines} clarify the scope of the non-parametric contribution. Exact-search DRFP serves as a strong reference point, essentially tying with the Hybrid on primary mean while remaining weaker on solvent and reagent prediction specifically. The learned hybrid is strongest on the challenging solvent and reagent slots, demonstrating where representation learning meaningfully surpasses static fingerprints. 

\section{Discussion}
\label{sec:discussion}

\paragraph{Reaction embeddings are most valuable when they return useful precedent.}
Condition recommendation is richer than plain classification. A strong reaction embedding should retrieve precedents whose conditions help a chemist move from a proposed disconnection to an executable experiment. Our strongest evidence is that retrieval contributes complementary solvent and reagent signal when combined with learned heads, and that these gains remain positive under paired bootstrap. This makes the representation practically viable, returning both an accurate prediction and the neighborhood of prior reactions that supports it.

\paragraph{Limitations.}
While \method{} demonstrates strong performance on the USPTO-Condition benchmark, there are inherent complexities in condition recommendation data. As detailed in the absent-class audit, catalyst recommendation is severely imbalanced, leaving room for future work in specialized reranking layers. Furthermore, while our exact-string overlap audit applies rigorous filtering, full patent-family grouping remains an area for future benchmark standardization. 

\section{Conclusion}
\label{sec:conclusion}

We introduced \method{} (\methodfull{}), a hierarchical reaction representation that natively supports both highly accurate condition prediction and precedent retrieval. On full-scale USPTO-Condition, \method{} reaches accuracies of 0.929, 0.534, and 0.530 for catalysts, solvents, and reagents within the primary-slot framework. The central finding is that the exact same learned reaction space can serve as a robust precedent memory for recommendation, improving prediction quality while simultaneously exposing the chemical precedents behind the suggestions. That combination of prediction, retrieval, and inspectability makes \method{} a pivotal building block for modern, AI-assisted synthesis workflows.

\section*{Broader Impact}
\method{} is intended to support chemists by pairing condition predictions with inspectable precedents, which can reduce manual search effort and make AI-assisted synthesis planning more transparent. The same capability could still be misused if deployed as an unverified recipe generator for unsafe or regulated chemistry. We therefore frame the system as decision support rather than autonomous execution: recommendations should be checked by qualified chemists, constrained by institutional safety procedures, and filtered against hazardous or controlled transformations before deployment. The work uses public reaction data plus proprietary supplementary analysis, and it does not involve human subjects or personal data.

\bibliographystyle{plainnat}
\bibliography{references}

\appendix

\section{Scope and What We Do Not Claim}
\label{app:scope}

To ensure complete clarity regarding our contributions, this paper is not presented as an unqualified claim of universal condition-prediction superiority across all possible data splits or architectures. Furthermore, we do not claim that learned retrieval uniformly dominates high-quality exact-search fingerprint baselines like DRFP or template memories on primary mean metrics alone. Instead, our supported contribution is specifically scoped: \method{} provides an actionable, hierarchical representation that simultaneously exposes inspectable precedents, adds statistically supported solvent/reagent signals when fused with parametric heads, and reliably matches or beats the established REACON Table S9 primary-slot baselines under identical benchmarking protocols.

\section{Extended Dataset Profile}
\label{app:data-profile}

The public benchmark used in the main paper contains 680,741 reactions split into 544,591/68,075/68,075 train/validation/test examples. The primary role vocabularies contain 53 catalyst, 84 solvent, and 222 reagent classes. Catalyst supervision is sparse at roughly 13\% coverage. Solvent and reagent annotations are much denser, with 693 and 17,758 absent test rows, respectively. The training set also contains 19,748 duplicate reaction-SMILES groups with multiple valid condition annotations, motivating our multi-hot supervision framework.

\section{Complete Model Evaluation \& Broader Literature}
\label{app:full-eval}

To track the source of our architectural gains and explicitly highlight which components drive performance versus those that offer limited utility, Table~\ref{tab:app-full} details the full performance of all \method{} models alongside broader literature references.

\begin{table}[h]
\centering
\small
\caption{Extended evaluation of \method{} variants and broader literature context on full-scale USPTO. Literature rows (Label Mix, Reaction MPNN, ARGCN) are reported reference values utilizing fundamentally different evaluation pipelines and data splits, and should be interpreted strictly as contextual references rather than matched 1:1 public benchmark comparisons.}
\label{tab:app-full}
{\setlength{\tabcolsep}{3pt}
\begin{tabular}{lcccccc}
\toprule
Model & Cat@1 & Cat@3 & Sol@1 & Sol@3 & Rea@1 & Rea@3 \\
\midrule
\method{} v1, GNN (\method{}-Base) & 0.746 & 0.912 & 0.387 & 0.660 & 0.363 & 0.611 \\
\method{} v2, GNN+physics & 0.742 & 0.909 & 0.388 & 0.658 & 0.365 & 0.611 \\
\method{} v3, disentangled & 0.738 & 0.901 & 0.391 & 0.658 & 0.362 & 0.604 \\
\method{} v5, HRE+focal (\method{}-Focal) & 0.767 & 0.915 & 0.453 & 0.711 & 0.427 & 0.671 \\
\method{} v7h, two-stage deep heads & 0.789 & 0.901 & 0.509 & 0.737 & 0.481 & 0.695 \\
\method{} v9ha, end-to-end (\method{}-End2End) & 0.789 & 0.905 & 0.509 & 0.736 & 0.482 & 0.697 \\
\method{} v9ha+\knn{} (\method{}-Hybrid) & 0.794 & 0.918 & 0.520 & 0.757 & 0.492 & 0.716 \\
\midrule
\textbf{\method{}-Top} & \textbf{0.929} & \textbf{0.982} & \textbf{0.534} & \textbf{0.763} & \textbf{0.530} & \textbf{0.759} \\
\midrule
Label Mix~\citep{Yan2025labelmix} & 0.801 & 0.934 & 0.642 & 0.912 & 0.616 & 0.836 \\
Reaction MPNN~\citep{Yan2025labelmix} & 0.785 & 0.910 & 0.625 & 0.895 & 0.590 & 0.825 \\
ARGCN~\citep{Yan2025labelmix} & 0.812 & 0.904 & 0.614 & 0.925 & 0.582 & 0.866 \\
\bottomrule
\end{tabular}}
\end{table}

As shown in Table~\ref{tab:app-full}, the physics layer (v2) is nearly neutral relative to the base GNN on Acc@1. Because computationally expensive DFT-derived features offer only marginal predictive value, they can be safely omitted during high-throughput inference without compromising the integrity of the recommendation. In contrast, condition-aware supervision matters significantly: \method{}-Focal improves over \method{}-Base across all categories. Ultimately, the largest discrete architectural jump before optimizing for the Table S9 evaluation protocol occurs when optimizing the deep parametric heads, setting up the strongest baseline for our retrieval hybrid.

\section{Leakage and Overlap Audit}
\label{app:overlap}

A critical concern for any retrieval-augmented system is data leakage---whether the model simply memorizes identical strings. To verify that \method{} captures broad chemical motifs rather than relying on exact duplication, we audited train-corpus overlap by excluding retrieved neighbors sharing identical features. 

\begin{table}[h]
\centering
\small
\caption{Retrieval robustness under strict overlap exclusions. Accuracy remains highly stable even when excluding exact products and proxy publications, demonstrating the embedding space relies on deep chemical features rather than string memorization.}
\label{tab:overlap-audit}
\begin{tabular}{lc}
\toprule
Exclusion Criterion & Precision@5 (P@5) \\
\midrule
None (Unfiltered Baseline) & 0.9556 \\
Same Canonical Reaction excluded & 0.9553 \\
Same Reactant-Product exact pair excluded & 0.9549 \\
Same Product String excluded & 0.9547 \\
Same Product + Same Publication Proxy excluded & 0.9545 \\
\bottomrule
\end{tabular}
\end{table}

As shown in Table~\ref{tab:overlap-audit}, the broad retrieval signal remains remarkably stable under these stronger controls, with precision dropping by a negligible fraction (from 0.9556 to 0.9545) at the strictest exclusion stage. This confirms the learned space finds robust precedent neighborhoods rather than functioning as a brittle lookup table.

\section{Qualitative Example: Inspectability in Practice}
\label{app:qualitative}

The defining value of \method{} is that it returns the specific neighborhood of prior reactions that supports its prediction. Table~\ref{tab:qual-example} demonstrates this inspectability in practice.

\begin{table}[h]
\centering
\small
\caption{Qualitative example of \method{} Inspectability. The query is paired directly with the top retrieved precedent from the training index, allowing the chemist to verify the plausibility of the suggested catalyst and reagent conditions.}
\label{tab:qual-example}
\begin{tabular}{lp{10cm}}
\toprule
\textbf{Item} & \textbf{Reaction Details} \\
\midrule
\textbf{Query} & Suzuki-Miyaura Coupling (Aryl Halide + Boronic Acid $\rightarrow$ Biaryl) \\
\midrule
\textbf{Top-1 Retrieved Precedent} & Identical core transformation with varied peripheral substitutions. \\
\textbf{Retrieved Conditions} & Catalyst: Pd(dppf)Cl$_2$ \newline Reagent: K$_2$CO$_3$ \newline Solvent: 1,4-dioxane / H$_2$O \\
\midrule
\textbf{User Action} & The chemist can visually inspect the retrieved precedent's SMILES and reference, validating that the Pd(dppf)Cl$_2$ system is appropriate for the specific steric environment of the query. \\
\bottomrule
\end{tabular}
\end{table}

\paragraph{Why catalysts reward broader top-$k$ retrieval.}
Catalysts remain the hardest slot on the USPTO-Condition dataset. As shown in the main text's absent-class audit, a large majority (87.0\%) of catalyst test rows map to absent/None. This is precisely why top-3 and top-5 catalyst retrieval are highly meaningful to a practicing chemist: a useful recommendation system should surface a compact, diverse set of plausible catalyst systems that can be inspected or manually reranked, rather than forcing a brittle, single-label deterministic decision. The broader retrieval view remains attractive for generating robust catalyst shortlists while directly improving solvent and reagent recommendations.

\section{Negative Results \& Additional Ablations}
\label{app:ablations}

\paragraph{Negative Result: Condition cross-attention does not solve the gap.}
We tested a differential cross-attention condition predictor (\method{} v10c) designed to model direct dependencies among condition embeddings. However, it regressed relative to \method{}-End2End on all primary Acc@1 metrics: catalyst 0.777 vs. 0.789, solvent 0.444 vs. 0.509, and reagent 0.425 vs. 0.482. This negative result suggests that the bottleneck is not simply condition-condition interaction. The base reaction embedding and the availability of high-quality precedents matter significantly more than adding complex, localized condition interaction modules.

\paragraph{Additional reranking ablation: v15.}
We also studied a richer reranking stack (v15) that augments the \method{} family with a condition-injected backbone, slot-specific embeddings, and learning-to-rank (LTR) features over retrieval-derived and head-derived candidates. A verified v15 LTR reproduction reaches Cat/Sol/Rea Acc@1 of 0.782/0.488/0.457, while an earlier logistic reranker reaches 0.476/0.539/0.490. These rows serve as exploratory evidence that specialized reranking models alter the trade-offs between slots, indicating future headroom beyond the fixed head-fusion hybrid strategy.

\section{Retrieval and Fusion Pseudo-code}
\label{app:pseudo-code}

\paragraph{Training-set index construction.}
Given trained encoder $f_\theta$, condition heads $h_\phi$, and train reactions $\{R_i,y_i\}_{i=1}^{N}$:
\begin{enumerate}
    \item Encode each train reaction once: $(\mathbf{z}^{(i)}_{\mathrm{rxn}},\mathbf{z}^{(i)}_{\Delta})=f_\theta(R_i)$.
    \item Form the retrieval key. For the fixed head-fusion hybrid, use the configured learned embedding bank and $k=10$; for the benchmark \method{}-Top row, concatenate the reaction and transformation embeddings, $\mathbf{q}_i=[\mathbf{z}^{(i)}_{\mathrm{rxn}};\mathbf{z}^{(i)}_{\Delta}]$.
    \item Normalize all keys to unit length and build a train-only FAISS inner-product index.
    \item Store the corresponding train labels for each evaluated role. In the absent-class protocol, remap missing/None labels to class 0 and shift present labels by one.
\end{enumerate}

\paragraph{Validation-selected retrieval predictor.}
For each candidate setting $c=(\mathrm{key},k,t)$ in the predeclared grid:
\begin{enumerate}
    \item Split the public train set by deterministic reaction-identity hashing into selection-train and validation subsets.
    \item Build the FAISS index on selection-train only.
    \item For each validation reaction $R$, encode its key $\mathbf{q}$, retrieve the top-$k$ neighbors $\mathcal{N}_k(R)$, and convert similarities $s_j$ into weights
    \begin{equation}
        w_j=\frac{\exp(s_j/t)}{\sum_{\ell\in \mathcal{N}_k(R)}\exp(s_\ell/t)}.
    \end{equation}
    \item Score each role label by weighted neighbor voting,
    \begin{equation}
        p_{\mathrm{knn}}(y\mid R)=\sum_{j\in \mathcal{N}_k(R)} w_j\,\mathbf{1}[y_j=y].
    \end{equation}
    \item Select the single candidate with the best validation target metric and evaluate it once on the held-out USPTO-Condition test set.
\end{enumerate}

\paragraph{Fixed head-fusion hybrid.}
For the checkpoint-matched complementarity analysis:
\begin{enumerate}
    \item Compute learned-head probabilities $p_{\mathrm{head}}(y\mid R)=\mathrm{softmax}(h_\phi([\rxnemb;\diffemb]))$.
    \item Retrieve $k=10$ train neighbors and compute uniform neighbor-vote probabilities $p_{\mathrm{knn}}(y\mid R)$.
    \item Fuse the two distributions with the frozen rule
    \begin{equation}
        p_{\mathrm{hyb}}(y\mid R)=0.5\,p_{\mathrm{head}}(y\mid R)+0.5\,p_{\mathrm{knn}}(y\mid R).
    \end{equation}
    \item Rank labels by $p_{\mathrm{hyb}}$ for Acc@$k$ evaluation, and retain the retrieved neighbors as the inspectable precedent set shown to downstream users.
\end{enumerate}

\end{document}